\documentclass[conference]{ieeeconf}

\IEEEoverridecommandlockouts
\usepackage[utf8]{inputenc}
\usepackage[english]{babel}

\usepackage{xspace}
\usepackage{graphicx}
\usepackage{wrapfig}

\usepackage{xcolor}
\definecolor{gk}{RGB}{120, 120, 120}
\definecolor{gg}{HTML}{5f9411}
\definecolor{gb}{HTML}{417598}
\definecolor{gr}{HTML}{d15120}
\definecolor{gy}{HTML}{d2ad00}

\usepackage{mathtools}

\usepackage{amsmath}
\usepackage{amssymb}
\usepackage{amsthm}

\usepackage[linesnumbered,ruled]{algorithm2e}
\SetArgSty{textnormal}

\usepackage{listings}

\newcommand{\mydots}{.\hspace{-.1em}.\hspace{-.1em}.}

\usepackage[
activate   = {true},
protrusion = false,
expansion  = true,
kerning    = true,
spacing    = true,
tracking   = false,
auto       = true,
selected   = true,
factor     = 1000,
stretch    = 15,
shrink     = 15,
]{microtype}

\makeatletter
\let\NAT@parse\undefined
\makeatother
\usepackage[pdfa,colorlinks,bookmarksopen,bookmarksnumbered,allcolors=gr]{hyperref}
\usepackage{cite}

\newcommand{\rrtstar}{\textsc{rrt*}\xspace}
\newcommand{\rrtconnect}{\textsc{rrt}Connect\xspace}

\newcommand{\nasa}{\textsc{nasa}\xspace}
\newcommand{\tmp}{\textsc{tamp}\xspace}
\newcommand{\io}{\textsc{io}\xspace}
\newcommand{\rtwo}{Robonaut~2\xspace}
\newcommand{\ompl}{\textsc{ompl}\xspace}
\newcommand{\ros}{\textsc{ros}\xspace}
\newcommand{\yaml}{\textsc{yaml}\xspace}
\newcommand{\dart}{\textsc{dart}\xspace}
\newcommand{\urdf}{\textsc{urdf}\xspace}
\newcommand{\srdf}{\textsc{srdf}\xspace}
\newcommand{\klampt}{\textsc{klamp't}\xspace}
\newcommand{\OpenRAVE}{\textsc{OpenRAVE}\xspace}
\newcommand{\moveit}{\textsc{MoveIt}\xspace}

\newcommand{\rbw}{\emph{Robowflex}\xspace}
\newcommand{\cpp}{\textsc{c++}\xspace}
\newcommand{\xml}{\textsc{xml}\xspace}
\newcommand{\api}{\textsc{api}\xspace}
\newcommand{\xacro}{\textsc{xacro}\xspace}
\newcommand{\movegroup}{\textsc{MoveGroup}\xspace}
\newcommand{\moveitcpp}{\emph{MoveIt}\textsc{cpp}\xspace}

\newcommand{\mref}[1]{\mbox{\autoref{#1}}\xspace}

\newcommand{\rbwurl}{\url{https://github.com/KavrakiLab/robowflex}}
\newcommand{\rbwdoc}{\url{https://kavrakilab.github.io/robowflex/}}

\title{\LARGE \bf
  \emph{Robowflex}: Robot Motion Planning with \emph{MoveIt} Made Easy
}

\author{%
  Zachary Kingston%
  \hspace{1em}Lydia E. Kavraki%
  \thanks{%
    Department of Computer Science, Rice University, Houston, TX, 77005, USA, {\tt \small \{zak,kavraki\}@rice.edu}.
    This work was supported in part by NSF 1718478, NSF 2008720, NSTRF 80NSSC17K0163, and Rice University Funds.
  }
}

\begin{document}

\maketitle
\thispagestyle{empty}
\pagestyle{empty}

\begin{abstract}
  \rbw is a software library for robot motion planning in industrial and research applications, leveraging the popular \moveit library and Robot Operating System (\ros) middleware.
  \rbw provides an augmented \api for crafting and manipulating motion planning queries within a single program, making motion planning with \moveit easy.
  \rbw's high-level \api simplifies many common use-cases while still providing low-level access to the \moveit library when needed.
  \rbw is particularly useful for 1) developing new motion planners, 2) evaluating motion planners, and 3) complex problems that use motion planning as a subroutine (e.g., task and motion planning).
  \rbw also provides visualization capabilities, integrations to other robotics libraries (e.g., \dart and Tesseract), and is complementary to other robotics packages.
  With our library, the user does not need to be an expert at \ros or \moveit to set up motion planning queries, extract information from results, and directly interface with a variety of software components.
  We demonstrate its efficacy through several example use-cases.
\end{abstract}

\section{Introduction}
\label{sec:introduction}


A core component of any autonomous system is \emph{motion planning}~\cite{Choset2005,LaValle2006,Kavraki2016}, which finds feasible motions that satisfy task requirements (e.g., reaching the goal, satisfying some motion constraint, etc.).
There are many motion planning software system for general manipulators;
a popular library for motion planning is \moveit~\cite{MoveIt}, which is built on top of the ubiquitous Robot Operating System (\ros) framework~\cite{Quigley2009}.
\moveit has four key advantages: it is widely adopted in industry and research, it is easy to setup for new robots and over 150 robots are already available~\cite{Coleman2014}, it is easy to integrate with a \ros system, and it has a large and vibrant open source community.
However, due to \moveit's massive scope and abstract architecture, many tasks are challenging for both engineers and researchers.
For example, it can be difficult to evaluate and develop planning algorithms, extend a planner's functionality, extract low-level information from planners, or use a planner within the scope of a broader planning algorithm, e.g., task and motion planning~\cite{Dantam2018}.

This paper introduces \rbw, a software library designed to simplify the use of \moveit for industrial and research applications of motion planning.
\rbw is a high-level \api to easily manipulate robots, collision environments, planning requests, and motion planners.
\rbw ``wraps'' the underlying \moveit library within a \cpp interface that provides many utilities that simplify the use and evaluation of motion planners.
Moreover, \rbw provides direct access to the implementation (that is, not through \ros messaging).
The key advantage of this approach is the ability to 1) develop self-contained scripts to evaluate motion planning, 2) retain the capability of easy system integration through \ros when necessary, and 3) implement integrated algorithms that use motion planning extensively (e.g., task and motion planning).
\rbw also integrates other libraries, e.g., the Open Motion Planning Library (\ompl)~\cite{Sucan2012}, \dart~\cite{Lee2018}, \ros Industrial's Tesseract~\cite{tesseract}, and visualization with Blender~\cite{blender}.
For example, these integrations enable comparison of \moveit's \rrtconnect against Tesseract's TrajOpt on the same scene in a single, short script.
We demonstrate the usefulness of \rbw in several example use-cases, such as benchmarking for motion planning experiments and an industry-focused use-case of evaluating \rtwo walking from \nasa.
\rbw is open-source\footnote{\rbwurl}, documented online\footnote{\rbwdoc}, and has already been used in a number of \mbox{publications~\cite{Chamzas2019,Chamzas2021, Pairet2021, Kingston2019, Quintero2021, Hernandez2019, Wells2021, Moll2021, Sobti2021,Pan2021, Chamzas2022}}.

\section{Background}
\label{sec:background}
\rbw is built around \moveit~\cite{MoveIt}, a widely-used\footnotemark{} software library designed to provide motion planning to \ros~\cite{Quigley2009} enabled robots.
\moveit has been successfully used with many robots, such as the PR2~\cite{Wyrobek2008}, Fetch~\cite{Wise2016}, and \nasa's \rtwo~\cite{Baker2017}.
\moveit also provides a setup assistant to easily configure new robots for motion planning~\cite{Coleman2014}.
\moveit provides default motion planning plugins such as sampling-based motion planning~\cite{Choset2005,LaValle2006,Kavraki2016} through \ompl~\cite{Sucan2012} and trajectory optimization (e.g.,~\cite{Zucker2013, Schulman2014}).

Typically\footnotemark{}, users interact with \moveit through the provided \movegroup program, which leverages \moveit's plugin-based architecture to provide a flexible, configurable motion planning service. 
While convenient for basic motion planning, \movegroup falls short when used outside this scope. 
For more advanced applications of motion planning, such as profiling or evaluating detailed aspects of a motion planner, changing parameters or components of a planner, or extracting more information from planners, it is insufficient.
To edit or improve the capabilities of \moveit, users must either edit or create plugin classes, which can be complicated for those who are unfamiliar with \moveit's internal structure.
Moreover, if users just want to use \moveit as a library, there are many parameters to load and configure (e.g., \urdf, \srdf, joint limits, kinematics plugins, planner configurations, etc.).
\rbw provides a \movegroup-like interface as a high-level \api, while also providing full access to underlying data, enabling access to these underlying structures to either modify or inspect\footnotemark{}.

\addtocounter{footnote}{-2}
\footnotetext{\url{https://moveit.ros.org/moveit/ros/2020/07/24/moveit-research-roundup.html}}
\addtocounter{footnote}{1}
\footnotetext{\url{https://ros-planning.github.io/moveit_tutorials/}}
\addtocounter{footnote}{1}
\footnotetext{\label{ft:mcpp}Note that \moveitcpp (\url{https://ros-planning.github.io/moveit_tutorials/doc/moveit_cpp/moveitcpp_tutorial.html}) and \url{https://github.com/PickNikRobotics/moveit_boilerplate} provide simple interfaces to load \moveit structures internally, but lack the abstraction, isolation, \io support, and support modules provided by \rbw.}

\rbw aims to support motion planning research, such as developing new \emph{sampling-based algorithms}~\cite{Choset2005,LaValle2006,Kavraki2016}.
A popular motion planning library used by many frameworks is the Open Motion Planning Library (\ompl)~\cite{Sucan2012}, and support of \ompl is paramount for many benchmarking applications~\cite{Moll2015}.
However, augmenting or customizing \ompl is difficult for real robotic systems, due to the difficulty of accessing the internals of the library through interfaces such as \moveit.
\rbw provides low-level access to \ompl, which is essential for rapid development and testing of novel motion planning techniques.
Moreover, it is possible to develop novel planners within the \rbw framework that are not tied to \moveit or \ompl.

Additionally, \rbw was designed to support algorithms that use motion planning as a subroutine.
For example, motion planning is used within \emph{task and motion planning} (\tmp), e.g.,~\cite{Kaelbling2011, Srivastava2014, Dantam2018}.
\tmp combines \textsc{ai} task planning with motion planning, generating a sequence of valid actions the robot should accomplish, each requiring a motion plan.
There has been work in integrating \tmp-like approaches in \moveit, e.g.,~\cite{Goerner2019}, however this approach only finds motions for predetermined sequences of actions.
Critically, \tmp requires both evaluating many motion plans to validate actions as well as using feedback from motion planning to inform task planning.
These requirements involve a deeper introspection into motion planning, which is at odds with the design of \movegroup.

Beyond \moveit, there are many excellent motion planning frameworks that are compatible with \ros through wrappers.
\OpenRAVE~\cite{Diankov2010} is a fully featured motion planning framework.
However, it is not actively developed and does not have the level of community support available in \moveit and \ros.
Similarly, \klampt~\cite{Hauser2016} is an all-in-one toolbox that specializes in handling contact, providing its own environment with \ros adapters.
The Robotics Library~\cite{Rickert2017} is an all-in-one library for real-time planning, but lacks the ability to connect to the broader robotics ecosystem through \ros.
As stated above, the benefit of using \moveit over the above libraries is that \moveit is easy to integrate with new robots, easy to integrate with \ros, and is actively developed with a large community.


\section{The \rbw Library}

\rbw is intended for use in research, education, and industry.
\rbw is informed by the following design goals.
\begin{enumerate}
\item \emph{Clarity of Interface}: Provide an easy to understand interface that meshes with intuitive understanding of the concepts involved in motion planning.
\item \emph{Minimize \ros}: Encapsulate \ros as much as possible such that it is easy to write independent programs.
\item \emph{Leverage \moveit's Ubiquity}: By being based on \moveit, many robots are supported in \rbw out of the box.
  The use of \moveit also enables \rbw scripts to effortlessly connect to the greater \ros system.
\item \emph{Unrestricted Access and Integration}: While providing a high-level \api, give access to underlying libraries so users are not hampered by \rbw in any way.
  A key advantage \rbw provides is that all the underlying data structures used in \moveit, \ompl, or other libraries can be accessed and modified. 
  \rbw provides a common interface and adapters to convert between the representations used by each library.
\item \emph{Consistency Across Versions}: \rbw provides adapters such that \rbw code is not tied to a specific version of \ros or \moveit.
  \rbw supports all versions from \ros Indigo to \ros Noetic.
\end{enumerate}

At a high-level, \rbw provides wrappers around core \moveit concepts so that it is easy to create and manage robots, scenes, and planners within a script.
The primary building blocks of \rbw are the robot's kinematics, the collision environment, and the motion planner.
Although other facades to \moveit exist, \rbw provides a higher level of abstraction that allows users who are unfamiliar with \moveit to still achieve complex programming tasks.
Moreover, these core features are supported by a suite of utilities, such as input-and-output helpers for a variety of formats, benchmarking tools, visualization within RViz, and more.
Beyond the core library are auxiliary modules for other robotics libraries, and support seamless integration between \rbw and native formats.

\subsection{Input and Output}
\label{sec:rbw:io}

Many complicated robotic problems require large amounts of file input and output (\io), e.g., for configuration, loading problems and scenes, and more. 
\rbw provides many \io helpers for common file-types used in robotics, such as:

\paragraph{Files and \ros Packages}
It is common to access files that exist in \ros packages that are specified with package \textsc{uri}s---\rbw provides helper functions to resolve file paths and to either open or create said file.

\paragraph{\xml Files}
Many configuration files in robotics are written in Extensible Markup Language (\xml) or in the macro \xml language \xacro.
\rbw provides helper functions to load these files (including unprocessed \xacro) as well as inject changes on the fly.

\paragraph{\yaml Files}
It is common to save and load \ros messages as \yaml files.
However, there is no easy way to load or save \ros messages in \yaml in \cpp.
\rbw has broad \yaml support, and can load \yaml from \ros Python and \ros topics.
Many \rbw classes can load \ros messages, e.g., motion planning requests, robot states, and more.
This makes it easy to load problems using \yaml files.

\paragraph{\ros Parameters}
Many \ros programs rely on the parameter server, a distributed key-value store available in \ros.
As a result, it is sometimes difficult to have multiple programs running simultaneously that require similar parameters, leading to issues with managing namespaces.
By default, \rbw uses an anonymous namespace so that many instances of \rbw code can simultaneously run.
Moreover, there is support to load \yaml files onto the parameter server, which is typically only available through \ros launch, making it easy to have scripts load their parameters.

There is also support for \ros bag files, \textsc{hdf5} files, various transformation representations, and others. 

\subsection{Robot Kinematics}
Commonly, the kinematics and geometry of a robot are described using the Universal Robot Description Format (\urdf).
Moreover, \moveit requires a Semantic Robot Description Format (\srdf) file, which describes additional properties such as what links are allowed to collide with each other, what groups of joints should be used for planning, and what parts of the robot are the end-effector.
There are also additional configuration files \moveit requires, such as \yaml files that describe kinematics plugins and additional joint limits.
Typically, most robots are configured through the \moveit setup assistant wizard which generates a \ros package containing all necessary configuration files.
Part of this configuration are complicated \ros launch files that contain parameters necessary to start \movegroup, and in order to modify the behavior of \moveit, both the program and launch file need to be modified.

\rbw takes care of all the necessary legwork to load and configure a robot without the need for a launch file.
Moreover, \rbw enables loading multiple robots within the same program, and duplicating robots if necessary---with default \moveit, this process can become convoluted.
An example of loading a robot is shown in~\mref{fig:rxcode0}.

\begin{figure}
  \vspace{0.6em}
  \begin{center}
\begin{lstlisting}[language=C++]
 namespace rx = robowflex;

 auto wam7 = //
   std::make_shared<rx::Robot>("wam7");
 wam7->initialize( //
   "package://barrett_model/robots/wam_7dof_wam_bhand.urdf.xacro",  // urdf
   "package://barrett_wam_moveit_config/config/wam7_hand.srdf",     // srdf
   "package://barrett_wam_moveit_config/config/joint_limits.yaml",  // joint limits
   "package://barrett_wam_moveit_config/config/kinematics.yaml"     // kinematics
 );
\end{lstlisting}
  \end{center}
  \vspace{-1.2em}
  \caption{%
    Loading a robot (here, a Barrett WAM\textregistered~arm) in \rbw.
    \vspace{-1.5em}
  }
  \label{fig:rxcode0}
\end{figure}

\subsection{Collision Environment}
The robot can be used to initialize a planning scene, which contains the collision geometry of the environment.
The scene can be used for adding and moving collision objects and computing collisions and distance to collision.
For example, scenes can be loaded from \yaml files, which encode full planning scenes:
\begin{lstlisting}[language=C++]
 auto scene = std::make_shared<rx::Scene>(wam7);
 scene->fromYAMLFile( //
  "package://robowflex_library/yaml/scene.yml");
\end{lstlisting}

Scenes also support adding collision objects programatically:
\begin{lstlisting}[language=C++]
 auto scene = std::make_shared<rx::Scene>(wam7);
 auto geometry = //
   rx::Geometry::makeCylinder(0.025, 0.1);
 auto pose = //
   rx::TF::createPoseXYZ( //
     -0.268, -0.826, 1.313,    // position
         0.,     0.,    0.));  // XYZ Euler
 scene->updateCollisionObject( //
   "cylinder", geometry, pose);
\end{lstlisting}
Note that many scenes can be loaded simultaneously, can be copied and modified, and saved and loaded to and from disk.

\subsection{Motion Planner}
\moveit uses a plugin-based system to load motion planning pipelines\footnote{\url{https://ros-planning.github.io/moveit_tutorials/doc/motion_planning_pipeline/motion_planning_pipeline_tutorial.html}}, which consist of adapters that filter and process both the planning request and output trajectory found by a motion planner.
\rbw provides an implementation to access any pipeline, and helpers for common plugins such as the default \ompl planning pipeline plugin.
To specify a planning request, a helper class is provided which simplifies the design of complex goal and path constraints, as well as setting start and goal states.
This helper class can also save and load motion planning requests to \yaml files, for later evaluation or setup.
Many planners can be loaded simultaneously and used in tandem.
Moreover, \rbw supports inserting new planners, either through \moveit's \api or its own.

\subsection{Example Script}
\label{sec:exscript}

\begin{figure}
  \vspace{0.6em}
  \begin{center}
\begin{lstlisting}[language=C++]
 // Create a default Fetch robot.
 auto fetch = std::make_shared<rx::FetchRobot>();|\label{line:robot}|
 fetch->initialize();
 
 // Create an empty scene.
 auto scene = std::make_shared<rx::Scene>(fetch);|\label{line:scene}|
 
 // Create the default planner for the Fetch.
 auto planner = // |\label{line:planner1}|
   std::make_shared< //
     rx::OMPL::FetchOMPLPipelinePlanner>(fetch);
 planner->initialize(); |\label{line:planner2}|
 
 // Create a motion planning request.
 rx::MotionRequestBuilder request( // |\label{line:request1}|
   planner, "arm_with_torso");

 // Set the start state.
 fetch->setGroupState("arm_with_torso",
   {0.05, 1.32, 1.40, -0.2, 1.72, 0, 1.66, 0});
 request.setStartConfiguration( //
   fetch->getScratchState());
 
 // Set the goal state.
 fetch->setGroupState("arm_with_torso",
   {0.27, 0.5, 1.28, -2.27, 2.24, -2.77, 1, -2});
 request.setGoalConfiguration( //
   fetch->getScratchState());
 
 // Set the desired planner.
 request.setConfig("RRTConnect"); |\label{line:request2}|
 
 // Do motion planning!
 auto result = planner->plan( //  |\label{line:planning}|
                 scene, request.getRequest());
\end{lstlisting}
  \end{center}
  \vspace{-1.2em}
  \caption{%
    A code snippet demonstrating basic motion planning on a Fetch robot~\cite{Wise2016}, from a ``stow'' position of the arm to an ``unfurled'' position.
    Note this does not use any \ros messages, similar to the internals of \moveit's \movegroup program.
    \vspace{-1.5em}
}
\label{fig:rxcode1}
\end{figure}

\mref{fig:rxcode1} shows a simple script for motion planning with \rbw using the Fetch robot.
The robot is loaded on line~\mref{line:robot}---\rbw comes with some preconfigured robots. 
An empty planning scene for the robot is created on line~\mref{line:scene}.
The standard \ompl planner for the Fetch is created and initialized in lines~\mref{line:planner1}~to~\mref{line:planner2}.
A simple request, which unfurls the Fetch's arm from the stow position to an extended position, is created in lines~\mref{line:request1}~to~\mref{line:request2}.
Finally, motion planning occurs on line~\mref{line:planning}.
A version of this script is available in the repository\footnote{\url{https://github.com/KavrakiLab/robowflex/blob/master/robowflex_library/scripts/fetch_test.cpp}}.

This simple script is akin to basic planning using the \moveit's \texttt{MoveGroupInterface} class\footnote{\url{https://ros-planning.github.io/moveit_tutorials/doc/move_group_interface/move_group_interface_tutorial.html}}.
The critical difference is that, rather than having to use \texttt{roslaunch} to run an instance of the \movegroup program and then communicate plans over \ros messages, all of \moveit's internal structures are loaded in the \rbw program%
\footnote{See \url{https://ros-planning.github.io/moveit_tutorials/doc/motion_planning_api/motion_planning_api_tutorial.html} for how this could be done without \rbw. Also, see \mref{ft:mcpp} for other helper classes that can set up planning without \rbw.}.
Providing access to these structures within a single program is a key benefit of \rbw.
The scripting paradigm offered by \rbw is more amenable to rapid testing and scripting than \movegroup, which is designed primarily to be a ``live'' component of a \ros system extant in some world.

\subsection{Integrations}

\rbw also provides a number of auxiliary modules that enable compatibility with different robotics libraries.

\paragraph{\ompl Integration}
The \rbw \ompl module provides deeper access for motion planning to the default \moveit \ompl motion planning plugin.
This includes extracting the underlying \ompl setup for a given planning problem, which enables users to modify the behavior of \ompl planning without having to recompile either the \moveit planning plugin or \ompl.
An example script for how to extract and customize the underlying \ompl planner used by \moveit is shown in~\mref{fig:rxcode3}.
With \rbw, it easy to use a custom planner or feature in \ompl, as compared to integrating a new planner into \moveit (e.g., this was done in~\cite{Chamzas2019, Chamzas2021, Pairet2021}).

\begin{figure}
  \vspace{0.6em}
  \begin{center}
\begin{lstlisting}[language=C++]
 // Create an OMPL planner
 auto planner = //
   std::make_shared< //
     rx::OMPL::OMPLInterfacePlanner>(fetch);

 // Extract underlying OMPL structures
 auto context = //
   planner->getPlanningContext(scene, request);
 auto ss = context->getOMPLSimpleSetup();

 // Customize OMPL planner
 ss->setPlanner(...);
 auto space = ss->getStateSpace();
 space->setStateSamplerAllocator(...);
\end{lstlisting}
\end{center}
\vspace{-1.2em}
\caption{%
  A code snippet demonstrating how to use the \rbw \ompl integration to access internal \ompl features.
  \vspace{-1em}
}
\label{fig:rxcode3}
\end{figure}

\paragraph{\dart Integration}
The \rbw \dart module provides an alternative to \moveit, by modeling robots and scenes in the \dart~\cite{Lee2018} framework with bidirectional conversion to/from \moveit constructs.
The module provides an implementation of motion planning through \ompl, including motion planning with manifold constraints~\cite{Kingston2019}.
Moreover, this module provides an easy way to plan for multi-robot systems, allowing arbitrary composition of \moveit enabled robots.
This capability was used by the multi-robot task-motion planning framework discussed in~\mref{sec:case:tmp}.
An example script demonstrating the \dart module is given shown in~\mref{fig:rxcode2}.

\begin{figure}[t!]
  \vspace{0.6em}
  \begin{center}
\begin{lstlisting}[language=C++]
 namespace rd = robowflex::darts;

 // Convert MoveIt robot to Dart
 auto fetch1 = std::make_shared<rd::Robot>(fetch);|\label{line:dconvert}|

 // Copy the Fetch for multi-robot planning
 auto fetch2 = fetch1->cloneRobot("other");|\label{line:dclone}|
 fetch2->setDof(4, 1); // Offset on X-axis

 // Combine kinematic structures into a world
 auto world = std::make_shared<rd::World>();
 world->addRobot(fetch1);
 world->addRobot(fetch2);

 // Plan using planning groups from both robots
 rd::PlanBuilder builder(world); |\label{line:dbuilder}|
 builder.addGroup("fetch", "arm_with_torso");
 builder.addGroup("other", "arm_with_torso");

 // Set start configuration for both robots
 builder.setStartConfiguration({ //
   0.05, 1.32, 1.4, -0.2, 1.72, 0, 1.66, 0,
   0.05, 1.32, 1.4, -0.2, 1.72, 0, 1.66, 0
 });
 builder.initialize();

 // Set goal configuration and setup planning
 auto goal = builder.setGoalConfiguration({ //
   0.27, 0.5, 1.28, -2.27, 2.24, -2.77, 1, -2,
   0.27, 0.5, 1.28, -2.27, 2.24, -2.77, 1, -2
 });
 builder.setGoal(goal);
 builder.setup();

 // Solve using OMPL
 auto result = builder.ss->solve(30.0);
\end{lstlisting}
\end{center}
\vspace{-1.2em}
\caption{%
  A code snippet demonstrating \rbw's \dart module for multi-robot motion planning.
  Here, the Fetch robot that was loaded in the prior script (\mref{fig:rxcode1}) is converted to \dart and copied, created a multi-robot system.
  A plan is generated in the composite space of both robots.
  \vspace{-1.5em}
}
\label{fig:rxcode2}
\end{figure} 

\begin{figure}
  \centering
  \includegraphics[width=0.95\linewidth]{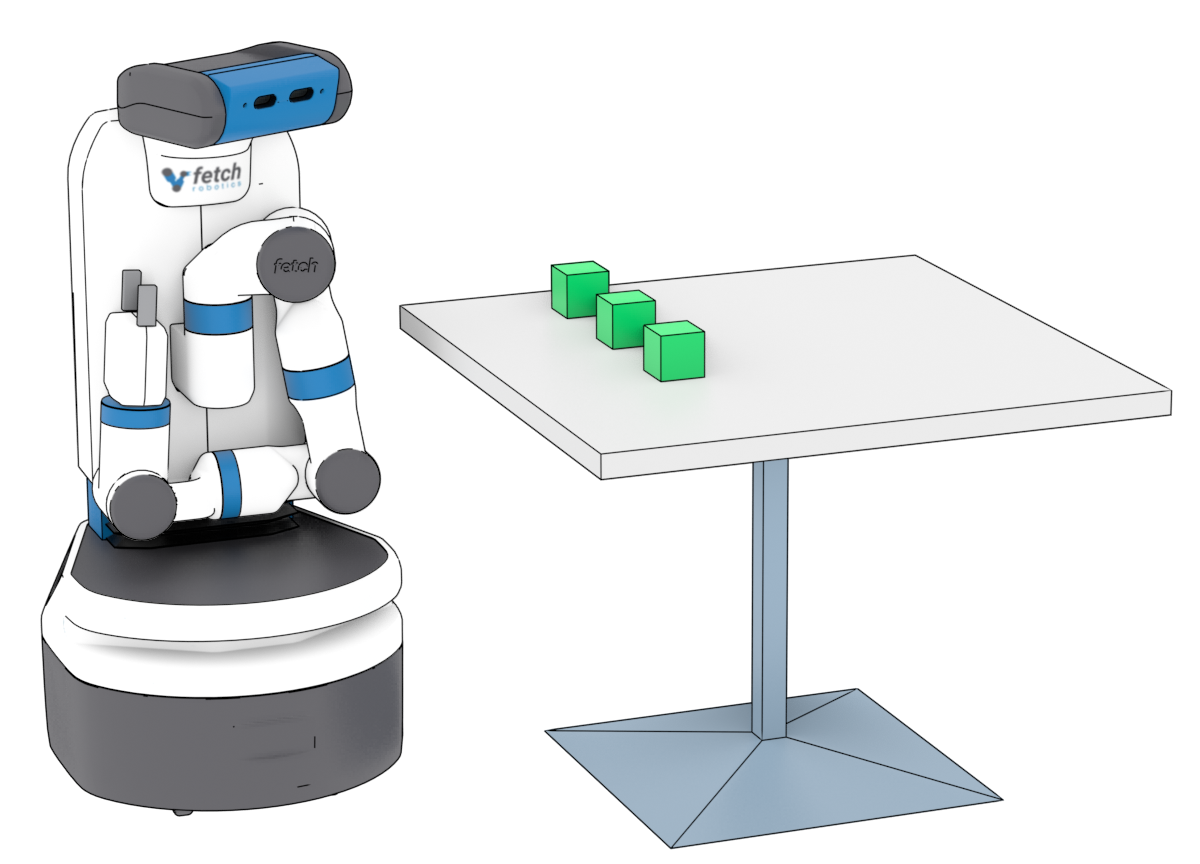}
  \caption{A Fetch robot~\cite{Wise2016} and planning scene rendered in Blender using \rbw's visualization module.}
  \label{fig:blender1}
\end{figure}

\paragraph{Tesseract Integration}
The \rbw Tesseract module provides access to the \ros Industrial Consortium's planning framework~\cite{tesseract}, which includes an implementation of the TrajOpt planner~\cite{Schulman2014}.
Similar to the \dart module, methods for converting data (e.g., scenes and plans) back and forth are provided.
\rbw's Tesseract module was used in~\cite{Quintero2021} to access and modify TrajOpt for comparison against \ompl planners.

\paragraph{\movegroup Integration}
The \rbw \movegroup module provides an easy connection to a live instance of \moveit's \movegroup.
This connection can be used to obtain the current collision environment, publish plans to be executed, and generally synchronize information between \movegroup and a \rbw script.

\paragraph{Visualization with Blender}

The \rbw visualization module makes it easy to render robots within Blender~\cite{blender}, a tool for 3D modeling and animation.
An example rendered still is shown in~\mref{fig:blender1}.
Moreover, it is easy to animate motion plans generated by \rbw to create appealing visualizations and videos\footnote{\url{https://kavrakilab.github.io/robowflex/md__home_runner_work_robowflex_robowflex_robowflex_visualization_README.html}}.
This module has also been used to generate figures in~\cite{Kingston2019, Chamzas2019, Chamzas2021}.

\section{Example Use-Cases}
\label{sec:case}

Crucial to the particular use-cases listed here as well as many others, \rbw simplifies the pipeline of creating, editing, inspecting, saving, and loading motion planning problems (both collision environments and motion planning requests).
These capabilities are essential for many research and industrial tasks that rely on reproducibility, experimentation, and debugging.

\subsection{Benchmarking Motion Planners}
\label{sec:case:bench}

\begin{figure}
  \centering
  \includegraphics[width=0.45\textwidth]{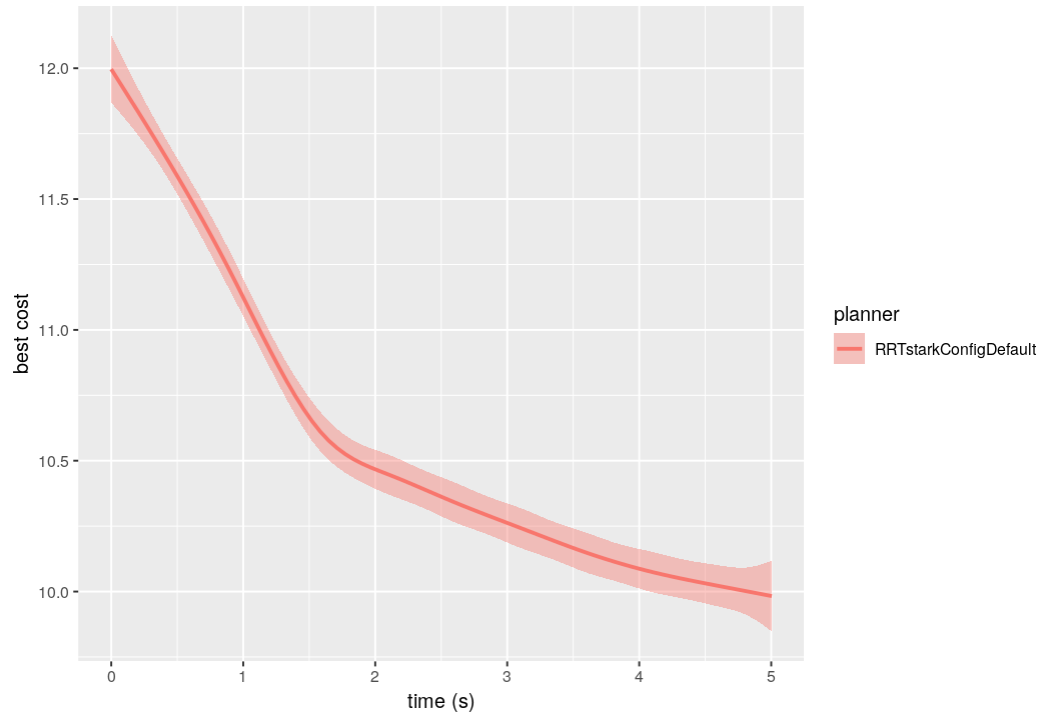}
  \caption{%
    Example benchmarking results from 200 runs of \rrtstar~\cite{Karaman2011}, an asymptotically optimal algorithm, on the Fetch.
    The ``best cost'' path (here, shortest path length) is displayed over time.
    This progress property of the planner is captured by \rbw's benchmarking, using the planner from the \ompl module.
    This plot was generated with Planner Arena\protect\footnotemark~\cite{Moll2015}, which accepts \ompl benchmark output for interactive plotting.
  }
  \label{fig:bench}
\end{figure}

A core use-case for \rbw is benchmarking motion planners in a variety of planning scenes.
The task of evaluating motion planners on realistic robots in many environments, extracting detailed planner information, and collating all collected data is difficult.
\rbw provides a benchmarking tool that enables easy benchmarking of different planners, scenes, and requests.
The tool enables configurable benchmark output in a number of formats.
For example, a default format provided is the \ompl benchmark log output, so output is compatible with the standard \ompl benchmarking suite and tools~\cite{Moll2015}.
In addition, recall that \rbw supports loading both environments and requests from disk, making it easy to craft datasets for evaluation.
As new planners are developed, targeted benchmarking can be done for many planner properties, a contribution to the motion planning community due to the difficulty of setting up consistent benchmarking criteria.
For example, the following could be added after line \mref{line:request2} of \mref{fig:rxcode1} to benchmark the motion planning request:
\begin{lstlisting}[language=C++]
  rx::Profiler::Options options;
  rx::Experiment experiment( //
    "example",  // Name of experiment
    options,    // Options for internal profiler
    60.0,       // Query timeout
    100);       // Number of trials
    
  experiment.addQuery("planner",
    scene, planner, request->getRequest());

  auto *dataset = experiment.benchmark();
  rx::OMPLPlanDataSetOutputter output( //
    "benchmark_example");
  output.dump(*dataset);
\end{lstlisting}

Moreover, benchmarking can capture \emph{progress properties} of a motion planner, if properly exposed.
These properties are important for profiling the performance of asymptotically \mbox{(near-)optimal} motion planning algorithms, such as \rrtstar~\cite{Karaman2011}.
An example is shown in~\mref{fig:bench}.
\rbw's benchmarking was used in~\cite{Kingston2019,Hernandez2019,Chamzas2019,Chamzas2021,Moll2021,Pairet2021,Quintero2021}.
\footnotetext{\url{http://plannerarena.org/}}

Compared to \moveit's built in benchmarking capabilities\footnote{\url{https://ros-planning.github.io/moveit_tutorials/doc/benchmarking/benchmarking_tutorial.html}},
\rbw provides a self-contained means of benchmarking that is more easily extendable.
\moveit's benchmarking requires use of \ros Warehouse for constructing benchmark datasets as opposed to \rbw's simple file storage, and does not easily support planning scenes with obstacle variation, which is important for learning-based methods.
For example, \cite{Chamzas2021,Moll2021} take advantage of these two features by running \rbw benchmarking instances in containers federated over many machines.
Moreover, \rbw enables creation of custom metrics with access to underlying planner results (e.g,. progress properties are not available in \moveit, properties of a particular method), and can be run as a single script.

\subsection{Task and Motion Planning}
\label{sec:case:tmp}

\begin{figure}
  \vspace{0.6em}
  \centering
  \includegraphics[width=0.9\linewidth]{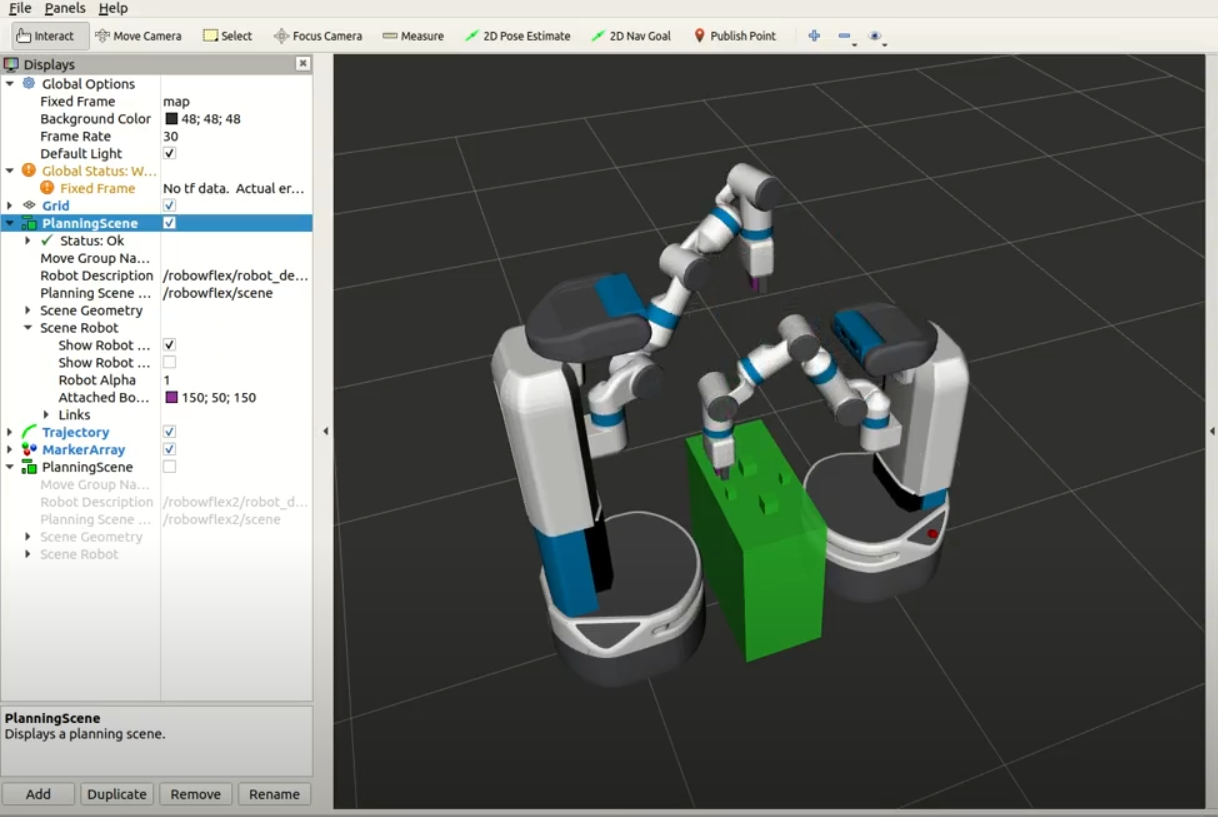}
  \caption{%
    Two Fetch robots~\cite{Wise2016} displayed in RViz executing a task and motion plan (\tmp) to rearrange blocks.
    The motion planning component of this \tmp problem is done through the \dart module of \rbw.
    \rbw enables the \tmp algorithm to have integrated access to the motion planner, which allows for trying many queries simultaneously, extracting collision information, and more.
    Image courtesy of Tianyang Pan.
  }
  \label{fig:mrtmps}
\end{figure}

One of the strengths of \rbw is motion planning in isolation.
That is, being able to use many different instances of robots, scenes, and motion planners all within the same script.
This is essential to efficient task and motion planning (\tmp), as a \tmp algorithm will evaluate many different motion plans in a variety of scenes to find a feasible task-and-motion plan.
\rbw has been used as the motion planning component in a few \tmp algorithms (e.g.,~\cite{Wells2021, Pan2021}), one of which is shown in~\mref{fig:mrtmps}.
Here, \rbw and \rbw's \dart module are leveraged to provide the motion planning components necessary for a multi-robot \tmp algorithm.
Crucial to \tmp is the evaluation of many possible scene configurations---\rbw allows for many copies of the collision environment to be considered in parallel.
Moreover, \rbw enables information on planning progress to be extracted from the underlying planner, which is used to inform the task planner.
Finally, the \dart module allows for multi-robot planning, as shown in~\mref{fig:rxcode2}.

\subsection{\rtwo and \nasa}
\label{sec:case:r2}

\begin{figure}
  \vspace{0.6em}
  \centering
  \includegraphics[width=0.9\linewidth]{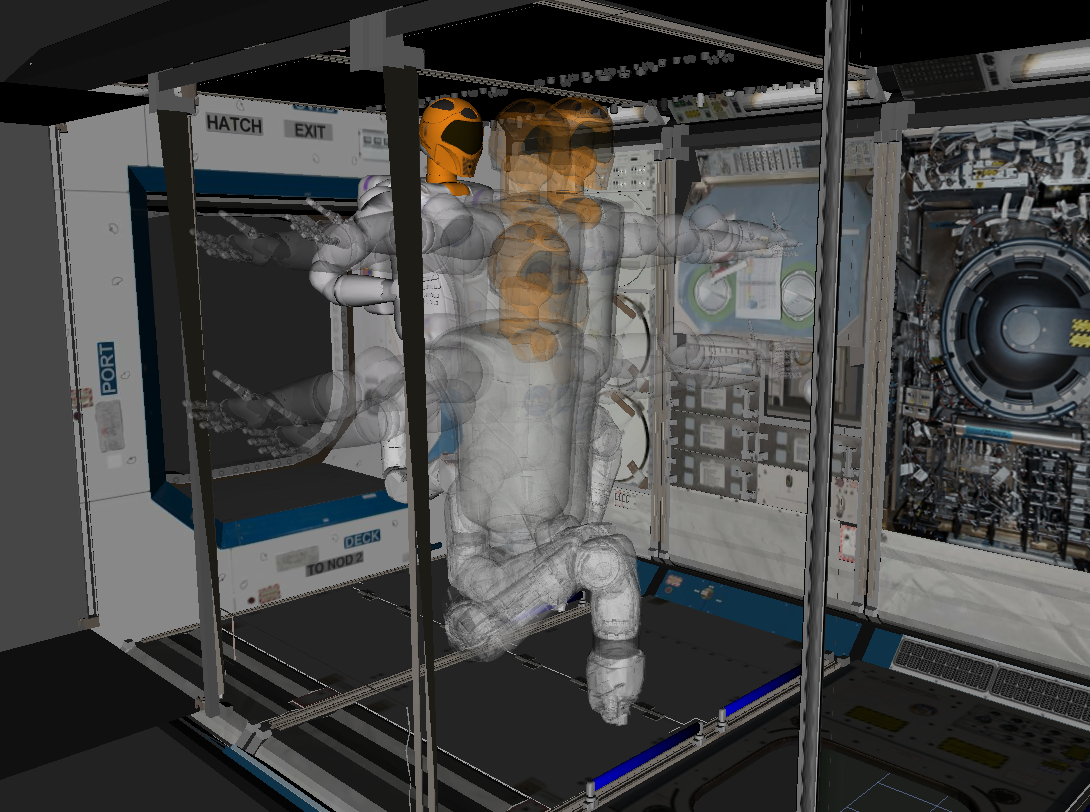}
  \caption{%
    \nasa's \rtwo inside of a module of the International Space Station, visualized in RViz.
    For a given step between handrails, many possible configurations are evaluated through \rbw.
    Image courtesy of Misha Savchenko and \nasa.
  }
  \label{fig:r2}
\end{figure}

\rbw has also been used by \nasa for motion planning for \rtwo.
\rtwo is a highly dexterous system, with many degrees of freedom.
One of the many motion planning challenges \rtwo faces is climbing across handrails in the International Space Station, as shown in~\mref{fig:r2}.
To this end, \rbw was used to evaluate potential handrail grasps and the difficulty of motion planning between different grasps to automate walking across the station.
\rbw provides the means to use, inspect, and evaluate custom inverse kinematics solvers for \rtwo and to benchmark the variety of handrail grasp configurations and scenes.
Additionally, \rbw was used for the \rtwo experiments and figure in~\cite{Kingston2019}.
As demonstrated by the examples presented here, the affordances provided by \rbw are general and broadly useful to different members of the robotics community.

\section{Discussion}
\label{sec:discussion}

We have presented \rbw, a \cpp library that enables the use of \moveit in an easier, more flexible way for the creation of advanced robot software for industry, research, and education.
The core advantage that \rbw provides over the default distribution of \moveit is the ability to easily access and modify core data structures within the program itself, rather than through \ros messages to the provided \movegroup program.
This also enables the use of motion planning within more complex algorithms, such as task and motion planning approaches.
Moreover, \rbw provides a high-level \api, enabling many use-cases such as benchmarking and motion planning without any \ros or \moveit expertise required.
\rbw also has a number of auxiliary modules that provide access to other robotics libraries and visualization tools, such as \ompl, \dart, Tesseract, and Blender.
Beyond motion planning software development, we hope that \rbw will enable broader application and adoption of motion planning algorithms and raise the level of experimental evaluation in comparisons.

\section*{Acknowledgments}
We thank Constantinos Chamzas, Carlos Quintero-Pe\~na, Andrew Wells, Wil Thomason, Bryce Willey, Juan David Hern\'andez, Mark Moll, and the other members of the Kavraki Lab. 
We also thank Misha Savchenko, Julia Badger, and the rest of the \rtwo team at \nasa Johnson Space Center. 

\bibliographystyle{IEEEtran}
\bibliography{references}

\end{document}